\documentclass[twoside,11pt]{article}

\usepackage{jmlr2e}
\usepackage{amsfonts}
\usepackage{amssymb}
\usepackage{amstext}
\usepackage{amsmath,bm}
\usepackage{natbib}
\usepackage{graphicx}
\usepackage{graphics}
\usepackage{booktabs}
\usepackage{array}
\usepackage{multirow}
\usepackage{algorithm}
\usepackage{algorithmic}
\usepackage{mathrsfs}
\usepackage{subfigure}
\usepackage{multicol}
\usepackage{makecell}
\usepackage{caption}

\usepackage{lastpage}
\jmlrheading{19}{2018}{1-\pageref{LastPage}}{12/17; Revised
10/18}{12/18}{17-772}{Jason Ge, Xingguo Li, Haoming Jiang, Han Liu, Tong Zhang, Mengdi Wang and Tuo Zhao}
\ShortHeadings{PICASSO for Sparse Learning in High Dimensions}{Ge, Li, Jiang, Liu, Zhang, Wang and Zhao}

\firstpageno{1}

\begin{document}

\title{\texttt{Picasso}: A Sparse Learning Library for High Dimensional Data Analysis in \texttt{R} and \texttt{Python}\thanks{Jason Ge and Xingguo Li contributed equally. Tuo Zhao is the corresponding author.}}

\author{\name Jason Ge$^{\dagger }$ \email jiange@princeton.edu
	\AND
        \name Xingguo Li$^{\S}$  \email lixx1661@umn.edu 
        \AND
        \name Haoming Jiang$^\diamond$ \email jianghm@gatech.edu
        \AND
        \name Han Liu$^\ddagger$ \email hanliu@northwestern.edu
        \AND
        \name Tong Zhang$^\ddagger$ \email  tongzhang@tongzhang-ml.org
        \AND
        \name Mengdi Wang$^\dagger$ \email mengdiw@princeton.edu 
        \AND
        \name Tuo Zhao$^\diamond$ \email tuo.zhao@isye.gatech.edu\\
        \addr{$\dagger$Department of Operations Research and Financial Engineering, Princeton University\\
        $^\S$Computer Science Department, Princeton University\\
        $^\diamond$School of Industrial and Systems Engineering, Georgia Tech\\
        $^\ddagger$Tencent Artificial Intelligence Lab
        }  
        }
        
        \vspace{-0.05in}

\editor{Cheng Soon Ong}

\vspace{-0.05in}

\maketitle

\begin{abstract}
We describe a new library named \texttt{picasso}
\footnote{More details can be found in our Github page: \url{https://github.com/jasonge27/picasso}},
 which implements a unified framework of pathwise coordinate optimization for a variety of sparse learning problems (e.g., sparse linear regression, sparse logistic regression, sparse Poisson regression and scaled sparse linear regression) combined with efficient active set selection strategies. Besides, the library allows users to choose different sparsity-inducing regularizers, including the convex $\ell_1$, nonvoncex MCP and SCAD regularizers. The library is coded in \texttt{C++} and has user-friendly \texttt{R} and \texttt{Python} wrappers. Numerical experiments demonstrate that \texttt{picasso} can scale up to large problems efficiently.
\end{abstract}


\section{Overview}

Sparse Learning arises due to the demand of analyzing high-dimensional data such as high-throughput genomic data \citep{neale2012patterns} and functional Magnetic Resonance Imaging \citep{liu2015calibrated}. The pathwise coordinate optimization is undoubtedly one the of the most popular solvers for a large variety of sparse learning problems. By leveraging the solution sparsity through a simple but elegant algorithmic structure, it significantly boosts the computational performance in practice \citep{friedman2007pathwise}. Some recent progresses in \citep{zhao2014general,ge2016homotopy} establish theoretical guarantees to further justify its computational and statistical superiority for both convex and nonvoncex sparse learning, which makes it even more attractive to practitioners.

We recently developed a new library named \texttt{picasso}, which implements a unified toolkit of pathwise coordinate optimization for solving a large class of convex and nonconvex regularized sparse learning problems. Efficient active set selection strategies are provided to guarantee superior statistical and computational preference. Specifically, we implement sparse linear regression, sparse logistic regression, sparse Poisson regression and scaled sparse linear regression \citep{tibshirani1996regression,belloni2011square,sun2012scaled}. The options of regularizers include the $\ell_1$, MCP, and SCAD regularizers \citep{fan2001variable, zhang2010nearly}. Unlike existing libraries implementing heuristic optimization algorithms such as \texttt{ncvreg} or \texttt{glmnet} \citep{breheny2013ncvreg,friedman2010regularization}, our implemented algorithm \texttt{picasso} have strong theoretical guarantees that it attains a global linear convergence to a unique sparse local optimum with optimal statistical properties (e.g. minimax optimality and oracle properties). See more details in \cite{zhao2014general, ge2016homotopy}.

\section{Algorithm Design and Implementation}

\begin{figure}[htb!]
	\begin{center}
		\includegraphics[width=0.7\linewidth]{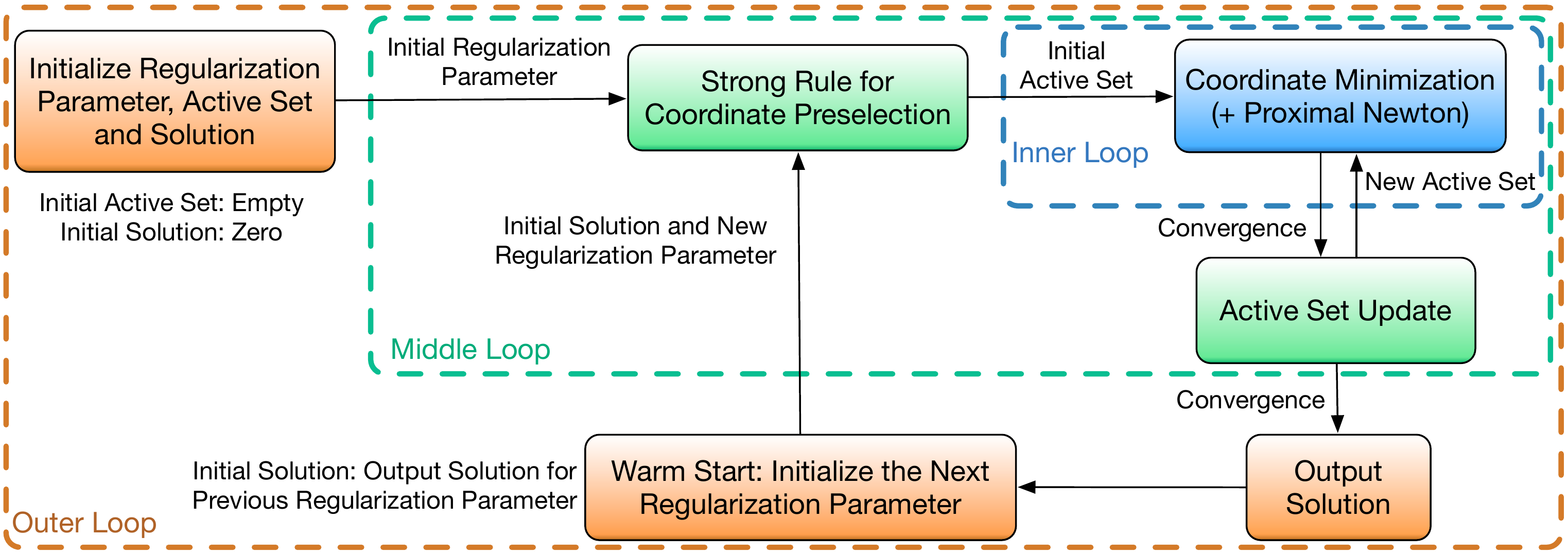}
					\vspace{-0.1in}
		\caption{The pathwise coordinate optimization framework with 3 nested loops: (1) Warm start initialization; (2) Active set selection, and strong rule for coordinate preselection; (3) Active coordinate minimization.}
		\label{fig:structure}
	\end{center}
\vspace{-0.45in}
\end{figure}

The algorithm implemented in \texttt{picasso} is mostly based on the generic pathwise coordinate optimization framework proposed by \cite{zhao2014general,ge2016homotopy}, which integrates the warm start initialization, active set selection strategy, and strong rule for coordinate preselection into the classical coordinate optimization. The algorithm contains three structurally nested loops as shown in Figure~\ref{fig:structure}:

{\bf \noindent (1)} Outer loop: The warm start initialization, also referred to as the pathwise optimization scheme, is applied to minimize the objective function in a multistage manner using a sequence of decreasing regularization parameters, which yields a sequence of solutions from sparse to dense. At each stage, the algorithm uses the solution from the previous stage as initialization. 

{\bf \noindent (2)}  Middle loop: The algorithm first divides all coordinates into active ones (active set) and inactive ones (inactive set) by a so-called strong rule based on coordinate gradient thresholding \citep{tibshirani2012strong}. Then the algorithm calls an inner loop to optimize the objective, and update the active set based on efficient active set selection strategies. Such a routine is repeated until the active set no longer changes

{\bf \noindent (3)} Inner loop: The algorithm conducts coordinate optimization (for sparse linear regression) or proximal Newton optimization combined with coordinate optimization (for sparse logistic regression, Possion regression, scaled sparse linear regression, sparse undirected graph estimation) only over active coordinates until convergence, with all inactive coordinates staying zero values. The active coordinates are updated efficiently using an efficient ``naive update'' rule that only operates on the non-zero coefficients. Better efficiency is achieved by the ``covariance update'' rule. See more details in \citep{friedman2010regularization}. The inner loop terminates when the successive descent is within a predefined numerical precision. 

The warm start initialization, active set selection strategies, and strong rule for coordinate preselection significantly boost the computational performance, making pathwise coordinate optimization one of the most important computational frameworks for sparse learning. The numerical evaluations show that \texttt{picasso} is highly scalable and efficient.

The library is implemented in \texttt{C++} with the memory optimized using sparse matrix output, and called from \texttt{R} and \texttt{Python} by user-friendly interfaces. Linear algebra is supported by the Eigen3 library \citep{eigenweb} for portable high performance computation. The implementation is modularized so that the algorithm in \texttt{src/solver/actnewton.cpp} works with popular sparsity-inducing regularizer functions and any convex objective function that exhibits restricted strong convexity property \citep{zhao2014general}. Users can easily extend the package by writing customized objective function subclass and regularizer function subclass following the virtual function interfaces of  \texttt{class ObjFunction} and \texttt{class RegFunction} in \texttt{include/picasso/objective.hpp}.

\vspace{-0.1in}

\section{Example of R User Interface} 

We illustrate the user interface by analyzing the eye disease data set in \texttt{picasso}.

\vspace{-0.1in}

\begin{verbatim}
> library(picasso); data(eyedata) # Load the data set
> out1 = picasso(x,y,method="l1",type.gaussian="naive",nlambda=20,
+                lambda.min.ratio=0.2) # Lasso
> out2 = picasso(x,y,method="mcp", gamma = 1.25, prec=1e-4)  # MCP regularizer
> plot(out1); plot(out2) # Plot solution paths

\end{verbatim}

\vspace{-0.28in}

The program automatically generates a sequence of regularization parameters and estimate the corresponding solution paths based on the $\ell_1$ and MCP regularizers respectively. For the $\ell_1$ regularizer, we set the number of regularization parameters as 20, and the minimum regularization parameter as \texttt{0.2*lambda.max}. For the MCP regularizer, we set the concavity parameter as $\gamma=1.25$, and the pre-defined accuracy as $10^{-4}$. Here \texttt{nlambda} and \texttt{lambda.min.ratio} are omitted, and therefore set by the default values (\texttt{nlambda=100} and \texttt{lambda.min.ratio=0.05}). We further plot two solution paths in Figure 2. 


%

\section{Numerical Simulation}


To demonstrate the superior efficiency of our library, we compare \texttt{picasso} with a popular \texttt{R} library \texttt{ncvreg} (version 3.9.1) for nonconvex regularized sparse regression, the most popular \texttt{R} library \texttt{glmnet} (version 2.0-13) for convex regularized sparse regression, and two \texttt{R} libraries \texttt{scalreg-v1.0} and \texttt{flare-v1.5.0} for scaled sparse linear regression. All experiments are evaluated on an Intel Core CPU i7-7700k 4.20GHz and under R version 3.4.3. Timings of the CPU execution are recored in seconds and averaged over 10 replications on a sequence of 100 regularization parameters. All algorithms are compared on the same regularization path and the convergence threshold are adjusted so that similar objective gaps are achieved.

\begin{figure}[h]
	\begin{minipage}[c]{0.5\textwidth}
		\footnotesize
		\centering
		\captionof{table}{Average timing  performance (in seconds) with standard errors in the parentheses and achieved objective values.}
		\scalebox{0.9}{
			\begin{tabular}[t]{l |@{ }c@{ }||@{}c@{}|@{ }c@{ }||@{}c@{}|@{}c}
				\Xhline{1 pt}
				\multicolumn{6}{c}{Sparse Linear Regression} \\
				\Xhline{1 pt}
				 \multicolumn{2}{c||}{ } & \multicolumn{2}{c||}{well-conditioned}  & \multicolumn{2}{c}{ill-conditioned}  \\
				\cline{1-6}
				\multirow{2}{*}{$\ell_1$}
				& \texttt{picasso} & 1.61(0.03)s  & 27.691  & 3.62(0.02)s & 32.543 \\
			 	& \texttt{glmnet} & 4.15(0.03)s  & 27.692  & 9.43(0.01)s & 32.537 \\
				& \texttt{ncvreg} & 5.92(0.01)s  & 27.690  & 6.66(0.01)s & 32.536 \\
				\cline{3-6}
				\multirow{2}{*}{SCAD} & \texttt{picasso} & 1.56(0.01)s  & 27.668  & 3.74(0.01)s & 33.133  \\
				& \texttt{ncvreg} &  5.61(0.01)s & 27.673  & 7.05(0.01)s & 33.156 \\
				\cline{3-6}
				\multirow{2}{*}{MCP} & \texttt{picasso} & 1.47(0.02)s  & 27.161  & 1.89(0.02)s & 32.468 \\
				& \texttt{ncvreg} & 4.07(0.03)s  &  27.161 & 2.56(0.01)s & 32.468 \\
				\cline{3-6}
				\Xhline{1pt}
				
				\multicolumn{6}{c}{Sparse Logistic Regression} \\
				\Xhline{1 pt}
				\multirow{2}{*}{$\ell_1$} & \texttt{picasso} 
				& 2.03(0.01)s  & 0.363  & 2.10(0.03)s & 0.327 \\
				& \texttt{glmnet} & 16.32(0.12)s  & 0.363  & 20.31(0.02)s & 0.327 \\
				& \texttt{ncvreg} & 4.04(0.01)s  & 0.363  & 62.89(0.04)s & 0.327 \\
				\cline{3-6}
				\multirow{2}{*}{SCAD} & \texttt{picasso} & 4.25(0.01)s  & 0.227  & 4.35(0.02)s  & 0.172  \\
				& \texttt{ncvreg} & 11.47(0.04)s  & 0.278  & error ($>$300s) &  \\
				\cline{3-6}
				\multirow{2}{*}{MCP} & \texttt{picasso} & 4.32(0.01)s  & 0.221  & 4.38(0.02)s & 0.165 \\
				& \texttt{ncvreg} & 9.37(0.08)s  & 0.248  & 6.99(0.01)s & 0.242 \\
				\cline{3-6}
				\Xhline{1 pt}
				\multicolumn{6}{c}{Scaled Lasso} \\
				\Xhline{1 pt}
				\multirow{2}{*}{$\ell_1$} &   \texttt{picasso} & 0.36(0.01)s  & 4.454  & 0.15(0.01)s & 5.495 \\
				&  \texttt{flare} & 5.23(0.06)s  & 5.188  & 297.36(2.77)s & 5.959 \\ 
				& \texttt{scalreg} & 40.20(0.63)s  & 4.492  & 49.12(10.98)s & 5.507 \\
				
				\cline{2-5}
				\Xhline{1pt}
				
				\Xhline{1pt}
			\end{tabular}
			\label{table:logit}
		}
	\end{minipage}
	\quad ~~\quad
	\begin{minipage}[c]{0.4\textwidth}
		\label{fig:path3}
		\begin{center}
			\begin{tabular}{c}
				\includegraphics[trim={0 0 0 0.3in},clip, width=0.85\linewidth]{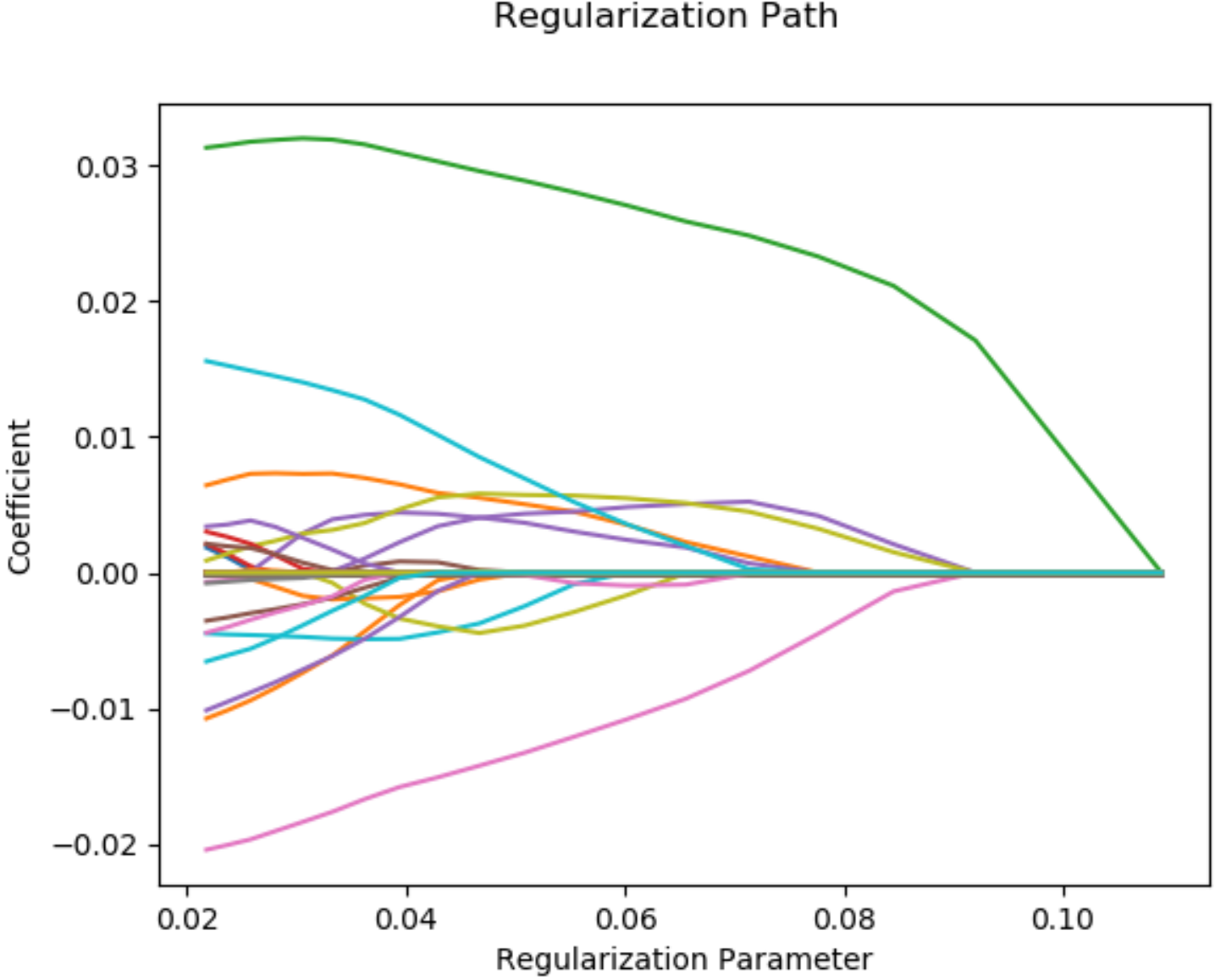} \\
				\includegraphics[trim={0 0 0 0.3in},clip, width=0.85\linewidth]{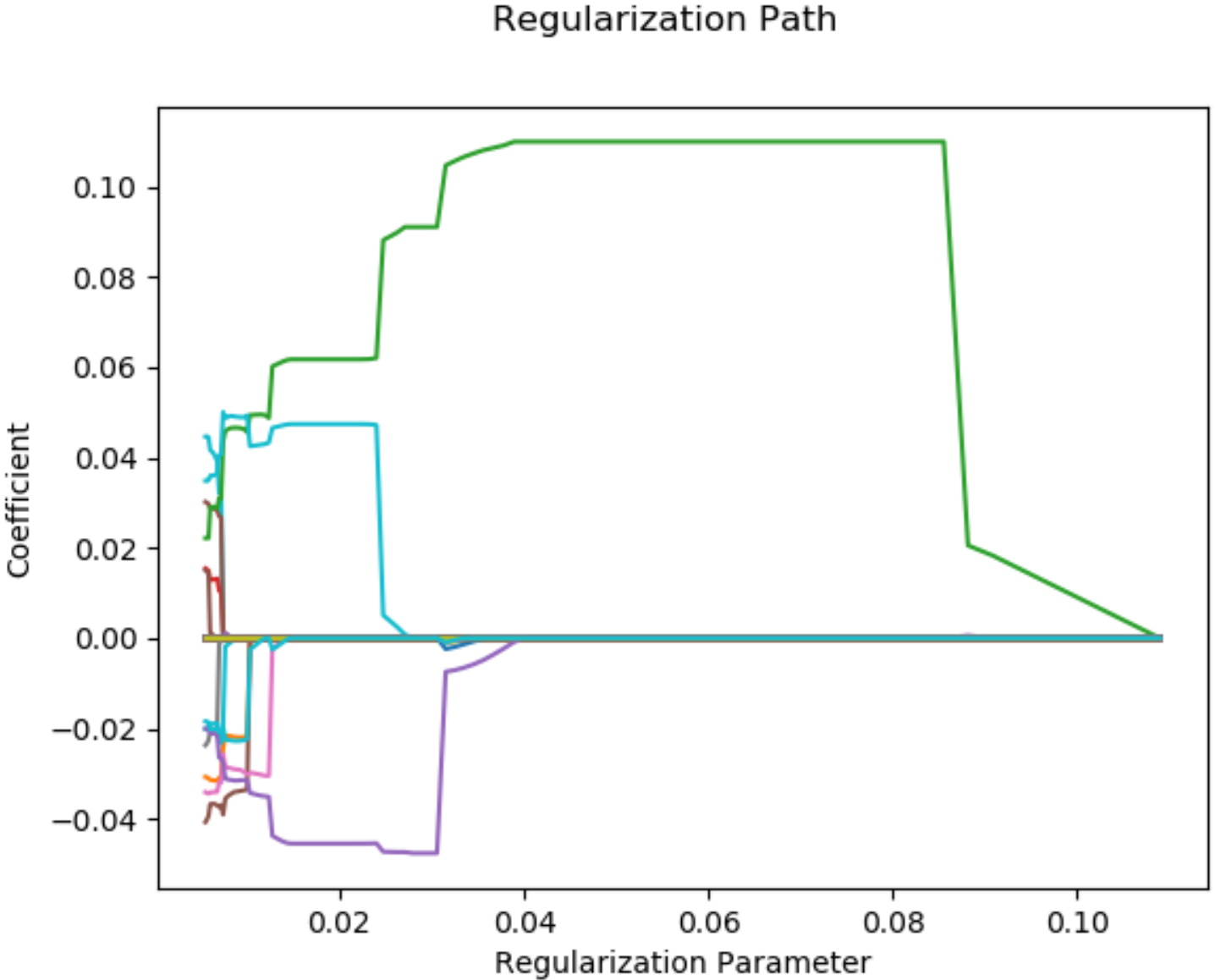}
			\end{tabular}  
		\end{center}
		\vspace{-0.25in}
		\captionof{figure}{The solution paths of $\ell_1$ (up) and MCP (down) regularizers.}
	\end{minipage}
\end{figure}

We compare the timing performance and the optimization performance in Table \ref{table:logit}. We choose the problem size to be $(n=3000, d=30000)$, where $n$ is the number of observation and $d$ is the dimension of the parameter vector. We tests the algorithms for both well-conditioned cases and ill-conditioned cases. The details of data generation can be found in the \texttt{R} library vignette. Here is our summary:

{\bf \noindent (1)} For sparse linear regression using any regularizer and sparse logistic regression using the $\ell_1$ regularizer, all libraries achieve almost identical optimization objective values, and \texttt{picasso} slightly outperforms \texttt{glmnet} and \texttt{ncvreg} in the timing performance.

{\bf \noindent (2)} For sparse logistic regression using nonconvex regularizers, \texttt{picasso} achieves comparable objective value with \texttt{ncvreg}, and significantly outperforms \texttt{ncvreg} in timing performance. 
We also remark that \texttt{picasso} performs stably for various settings and tuning parameters. However, \texttt{ncvreg} may converge very slow or fail to converge for sparse logistic regression using nonconvex regularizers, especially when the tuning parameters are relatively small (corresponding to denser estimators), as the ill-conditioned SCAD case shows. 

{\bf \noindent (3)} For scaled Lasso, in order to make other competitors (\texttt{flare} and \texttt{scalreg}) converges in resonable time, we swtich to a smaller problem size $(n=1000,d=10000)$. We see that \texttt{picasso} much more time saving than \texttt{flare} and \texttt{scalreg}.

\vspace{-0.075in}

\vspace{-0.2in}
\section{Conclusion}

The \texttt{picasso} library demonstrates significantly improved computational and statistical performance over existing libraries for nonconvex regularized sparse learning such as \texttt{ncvreg}. Besides, \texttt{picasso} also shows improvement over the popular libraries for convex regularized sparse learning such as \texttt{glmnet}. Overall, the \texttt{picasso} library has the potential to serve as a powerful toolbox for high dimensional sparse learning. We will continue to maintain and support this library.

\bibliography{Library}

\end{document}